\documentclass[a4paper]{article}

\usepackage{times}
\usepackage{graphicx} % more modern
\usepackage{subfigure} 

\usepackage{natbib}

\usepackage{algorithm}
\usepackage{algorithmic}
\usepackage[accepted]{icml2017}

\usepackage{float}
\usepackage{color}

\usepackage[english]{babel}
\usepackage[utf8x]{inputenc}
\usepackage[T1]{fontenc}

\usepackage[a4paper,top=3cm,bottom=2cm,left=3cm,right=3cm,marginparwidth=1.75cm]{geometry}

\usepackage{amsmath,amssymb}
\usepackage{graphicx}
\usepackage[colorinlistoftodos]{todonotes}
\begin{document} 
%\title{A Probe into Understanding GAN and VAE models}
%\author{
%{Jingzhao Zhang\thanks{Equal Contribution}\space , \space Lu Mi\printfnsymbol{1} }\\
%Massachusetts Institute of Technology, Cambridge, USA\\

%}
\twocolumn[
\icmltitle{A Probe Towards Understanding GAN and VAE Models}
\icmlsetsymbol{equal}{*}

\begin{icmlauthorlist}
\icmlauthor{Lu Mi}{mitcsail,equal}
\icmlauthor{Macheng Shen}{mitlids,equal}
\icmlauthor{Jingzhao Zhang}{mitlids,equal}\\
\enspace
\icmlauthor{\small{$^1$MIT CSAIL, $^2$MIT LIDS}}{}
\\
\icmlauthor{\small{\{lumi, macshen, jzhzhang\}@mit.edu}}{}
\\

\icmlauthor{\small{$*$ The authors equally contributed to this work.}}{}
\\
\icmlauthor{\small{$\dagger$ This report was a part of the class project for 6.867 Fall 2017. }}{}
\\
\end{icmlauthorlist}
\icmlaffiliation{mitcsail}{MIT CSAIL}
\icmlaffiliation{mitlids}{MIT LIDS}
\icmlcorrespondingauthor{Lu Mi}{lumi@mit.edu}
\icmlcorrespondingauthor{Macheng Shen}{macshen@mit.edu}
\icmlcorrespondingauthor{Jingzhao Zhang}{jzhzhang@mit.edu}
\vskip 0.3in

]

%\printAffiliationsAndNotice{\icmlEqualContribution} 

\begin{abstract} 
This project report compares some known GAN and VAE models proposed prior to 2017. There has been significant progress after we finished this report. We upload this report as an introduction to generative models and provide some personal interpretations supported by empirical evidence. Both generative adversarial network models and variational autoencoders have been widely used to approximate probability distributions of data sets. Although they both use parametrized distributions to approximate the underlying data distribution, whose exact inference is intractable, their behaviors are very different. We summarize our experiment results that compare these two categories of models in terms of fidelity and mode collapse. We provide a hypothesis to explain their different behaviors and propose a new model based on this hypothesis. We further tested our proposed model on MNIST dataset and CelebA dataset. 
\end{abstract} 

% Comments on Experiment: 1. Compare entropy change for GAN, WGAN, VAE in the same format. (Put GAN, WGAN, VAE into the same figure.)
% 2. Remember that in addition to entropy, we also want to compare the accuracy (otherwise, Gaussian noise image will also have high entropy). Add another figure for the last iteration sample for all experiments and discuss the image quality for each model. How much do they resemble samples from the original dataset?
% 3. Combine GAN WGAN VAE in experiment section into a single subsection named experiment setup. Systematically explain experiment environment, model structure, training parameter, etc, so that it provides enough information for other people to reproduce the result.

\section{Introduction} 
One way to interpret the goal of unsupervised learning algorithms \cite{lda,inference,mcmc} is that they try to describe the distribution of the true data using samples from the dataset. These algorithms all model the dataset with some probability distribution, and learn an approximate distribution from the data samples. However, without further constraints, solving such problems in high dimension is intractable, which requires exponentially growing number of samples. As a result, practical algorithms balance the model complexity and sample complexity to trade off model accuracy for efficiency.

More formally, assuming that random samples from a data set are drawn from an underlying true distribution $X \sim p(x)$, our goal is to design an algorithm that produces a distribution $q(x)$ based on i.i.d samples ${x_1, x_2,..., x_n}$ from the true distribution. Mathematically, the algorithm aims to minimize the divergence 
$$D_{\phi}(p|q) = E_p[\phi(\frac{q(x)}{p(x)})],$$ where the function $\phi$ is determined by the actual application.

For most function $\phi$ (e.g. $\phi = \log$ in KL-divergence), the divergence is minimized when $q(x) = p(x)$ almost everywhere. However, finding such a distribution $q(x)$ exactly requires infinite number of samples. Therefore, many algorithms parametrize the approximate distribution by $q(x; \theta)$, such that it only searches within a probability family $Q = {q(x; \theta)| \theta \in \Theta}$ of nice properties that make solving the problem more tractable. The latent Dirichlet analysis, for example, uses conjugate prior distributions in graphical models to allow deriving analytic expression of maximum likelihood estimators. In addition, parameterizing builds prior information into the model as a regularization and leads to better generalization results.

However, the problem of finding a good distribution family $Q$ itself may be hard. When $Q$ is too general, the algorithm may consume too many samples or require too much computing power. When $Q$ is not expressive enough, the result may have a large bias. In order to solve this problem, some works \cite{networkflow,hierarchy,entropy} proposed to use neural networks to parametrize probability distribution. The high representation ability of neural networks along with backpropagation algorithms make these algorithms very generalizable and efficient.

In addition to these results, another line of works \cite{gan,vae,wgan,vaegan,cgan,dcgan} is more dedicated to approximating data distribution in image datasets. These models generate high quality natural images and hence have attracted much attention in recent years. However, it is also widely known that GAN style models are very sensitive to training parameter tuning and suffers from unstable convergence and mode collapse. In our project, we first provide a brief overview of these models in section 2. We then reproduce the experiments of four different generative models and compare their performance in terms of image diversity (measured by entropy) and image fidelity in section 3. Based on these results, we propose a hypothesis that explains the difference between GAN and VAE in section 4. We further propose a new model and test on MNIST and CelebA datasets. The experiment results are also included in this project.

\section{Deep generative models}
Generative adversarial network (GAN) and Variational autoencoder (VAE) are two commonly used deep generative models that can generate complicated synthetic images. In this section, we will introduce four variations of GAN and VAE: (1) Vanilla GAN \cite{gan}, (2) Wasserstein GAN (WGAN) \cite{wgan}, (3) Vanilla VAE \cite{vae}, (4) VAE-GAN \cite{vaegan}. We will focus on the intuition, mathematical formulation and the issues with each of the models.

\subsection{Generative adversarial network}
GAN \cite{gan} uses two deep neural networks (namely, a generator and a discriminator) to train a generator of images. The generator is typically a de-convolutional neural network (DCN), and the discriminator is typically a convolutional neural network (CNN). During training, the generator takes in fixed dimensional noise vectors, which are called the latent variable, and outputs images. The generated synthetic images are blended with the true images from a dataset and fed into the discriminator. The classification accuracy of the discriminator is then fed back to the generator. Therefore, the training objective of the generator is to increase the classification error of the discriminator and that of the discriminator is to decrease the classification error. This training objective can be summarized as the following minimax problem:
\begin{equation}
\begin{aligned}
\min_{G}\max_{D}V(D,G)=&\mathbb{E}_{\bf{x}\sim p_{data}(\bf{x})}[log D(\bf{x})]\\
&+\mathbb{E}_{\bf{z}\sim p_{\bf{z}}(\bf{z})}[1-log D(G(\bf{z}))],
\end{aligned}
\label{gan_loss1}
\end{equation}
where $G$ is the mapping from the latent space to the data space, and $D$ is the discriminator loss measuring how well the discriminator classifies the blended data.\\ 
\indent The optimization problem defined by Eq. \ref{gan_loss1} can be viewed as a zero-sum game, which is shown to have a unique equilibrium point. This equilibrium point corresponds to the optimal distribution of the generated image, induced by the generative network, that solves the optimization problem. This provides a general framework for training of deep generative models. Nonetheless, it turns out that the training of this model is difficult when the discriminator is trained too well. That is, if the discriminator is too powerful, then the training gradient for the generator will vanish. Therefore, the authors of the GAN paper proposed another loss function:
\begin{equation}
\begin{aligned}
\min_{G}\max_{D}V(D,G)=&\mathbb{E}_{\bf{x}\sim p_{data}(\bf{x})}[log D(\bf{x})]\\
&-\mathbb{E}_{\bf{z}\sim p_{\bf{z}}(\bf{z})}[log D(G(\bf{z}))].
\end{aligned}
\label{gan_loss2}
\end{equation}
The problem with this optimization is that the resulted optimal distribution suffers from mode collapse. That is, the optimal distribution can only represent a sub-class of instances appearing in the data distribution. It turns out that both of the training difficulty and the mode collapse problem are due to the inappropriate functional form of the loss function. This is modified in WGAN such that these two problems are avoided.
\subsection{Wasserstein GAN}
It has been shown in \cite{wgan} that the first optimization, Eq. \ref{gan_loss1}, is essentially equivalent to minimizing the following objective, when the discriminator is fixed and optimal:
\begin{equation}
2JS(P_{data}||P_G)-2log 2,
\end{equation}
where $JS$ is the Jensen–Shannon divergence. When $P_{data}$ and $P_G$ are quite different from each other, $JS(P_{data}||P_G)$ becomes a constant. Therefore, the gradient vanishes, which is problematic for training with gradient descent.\\
\indent Likewise, the second optimization, Eq. \ref{gan_loss2}, is essentially equivalent to minimizing the following objective, when the discriminator is fixed and optimal:
\begin{equation}
KL(P_G||P_{data})-2JS(P_{data}||P_G),
\end{equation}

where $P_G$ is the distribution of the generator, $P_{data}$ is the data distribution, $KL$ is the Kullback–Leibler divergence. This is undesirable, as it wants to minimize the KL divergence while maximize the JS divergence simultaneously, which does not make sense. \\
\indent Moreover, this objective function assigns different penalty to two different types of error that the generator makes. Suppose $P_G(x)\rightarrow 0, P_{data}(x)\rightarrow 1$, which means the generator does not generate a realistic image, the corresponding penalty $KL(P_G||P_{data})\rightarrow 0$. However, suppose $P_G(x)\rightarrow 1, P_{data}(x)\rightarrow 0$, which means the generator generates images that do not look like those in the data, the corresponding penalty $KL(P_G||P_{data})\rightarrow +\infty$. Therefore, this loss encourages generating replicated images that have low penalty rather than generating diverse data that could result in a high penalty, thus causing mode collapse.\\
\indent WGAN solves the training difficulty and the mode collapse problem by using a modified loss function shown in Eq. \ref{wgan}, which essentially corresponds to minimizing the Wasserstein distance between the generative distribution and the data distribution. The W-distance has nice property that even two distributions have little overlap, the W-distance still varies smoothly.\\
\begin{equation}
\begin{aligned}
\min_{G}\max_{D}V(D,G)=&\mathbb{E}_{\bf{x}\sim p_{data}(\bf{x})}[D(\bf{x})]\\
&+\mathbb{E}_{\bf{z}\sim p_{\bf{z}}(\bf{z})}[1-D(G(\bf{z}))],
\end{aligned}
\label{wgan}
\end{equation}
\indent Although WGAN avoids mode collapse, we found that the generated images still do not look very realistic, as there is no term in the objective function that encourages the synthetic data to look like the training data. This is encouraged implicitly in another type of deep generative model, VAE.
\subsection{Variational autoencoder}
The idea behind VAE is to use a generative neural network and a recognition neural network to solve the variational inference problem that maximizes the marginalized data likelihood. The generative network obtained at the end of this process can generate synthetic data that looks similar to the training data. Nonetheless, the exact data likelihood is not easily obtainable, thus VAE approximately maximizes the evidence lower bound (ELBO) by gradient ascent on the following objective function:
\begin{equation}
\begin{aligned}
\mathcal{L}(\bf{\theta},\bf{\phi};\bf{x})=&\sum_{i=1}^{N}-KL(q_{\bf{\phi}}(\bf{z}|\bf{x}^{(i)})|| p_{\bf{\theta}}(\bf{z}))\\
&\mathbb{E}_{q_{\bf{\phi}}(\bf{z}|\bf{x}^{(i)})}[log \mbox{ }p_{\bf{\theta}}(\bf{x}^{(i)}|\bf{z})],
\end{aligned}
\label{vae}
\end{equation}
where $\bf{\theta}$ is the parameter of the generative network, and $\bf{\phi}$ is the parameter of the recognition network. $p_{\bf{\theta}}(\bf{z})$ is the distribution of the latent variable $\bf{z}$, which is represented by a Gaussian distribution whose mean and covariance is obtained by passing a noise parameter $\epsilon$ through the generative network. $q_{\bf{\phi}}(\bf{z}|\bf{x}^{(i)})$ is the approximate posterior distribution of the latent variable conditioned on the data instance, which is approximated as a Gaussian whose mean and covariance are obtained by passing the data instances through the recognition network. By minimizing the KL divergence between these two distributions, the model encourages the generated data to look similar to the training data. On the other hand, the second term in the objective function encourages the generative distribution $p_{\bf{\theta}}(\bf{x}^{(i)}|\bf{z})$ to be as diffusive as possible. Therefore, the resulted synthetic image is blurred, which is not desirable.
\subsection{VAE-GAN}

\begin{figure}[ht]
\vskip 0.2in
\begin{center}
\centerline{\includegraphics[width=\columnwidth]{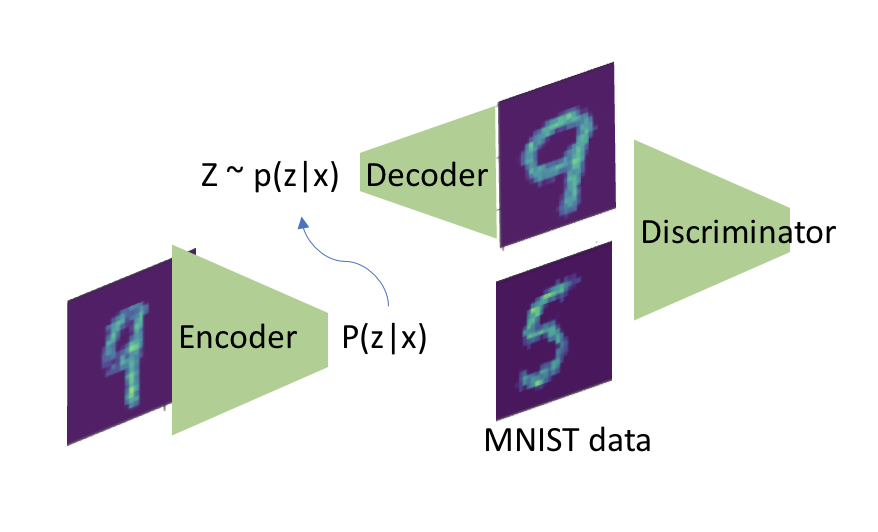}}
\caption{The architecture of the VAE-GAN model.}\label{vaegan}
\end{center}
\vskip -0.2in
\end{figure} 

One advantage of VAE models over GAN models is that it could map an input in the original dataset to latent factors and further to an image in the generator's approximation. However, the sample images generated by VAE are usually blurry and of lower quality compared to those from GAN models. To get the benefits of both, the work \cite{vaegan} proposed an architecture shown in fig.\ref{vaegan}, on top of the original VAE models. The VAE-GAN model adds a discriminator on top of the generated image. The loss function for the discriminator is the same as the one in GAN. The loss for the decoder and the encoder have two components. The first component is the same as Eq.\ref{vae} from VAE. The second component is 
\begin{equation}
 L_{GAN} = -\mathbb{E}_{\bf{z}\sim p_{\bf{z}}(\bf{z})}[log D(G(\bf{z}))].
\label{vae-ganloss} 
\end{equation}
This component is minimized when the generator successfully fools the discriminator. With this architecture, the VAE-GAN successfully generates GAN-style images while preserving the functionality to map a sample image back to its latent variables.

\begin{figure*}[ht]
\vskip 0.2in
\begin{center}
\centerline{\includegraphics[width=\linewidth]{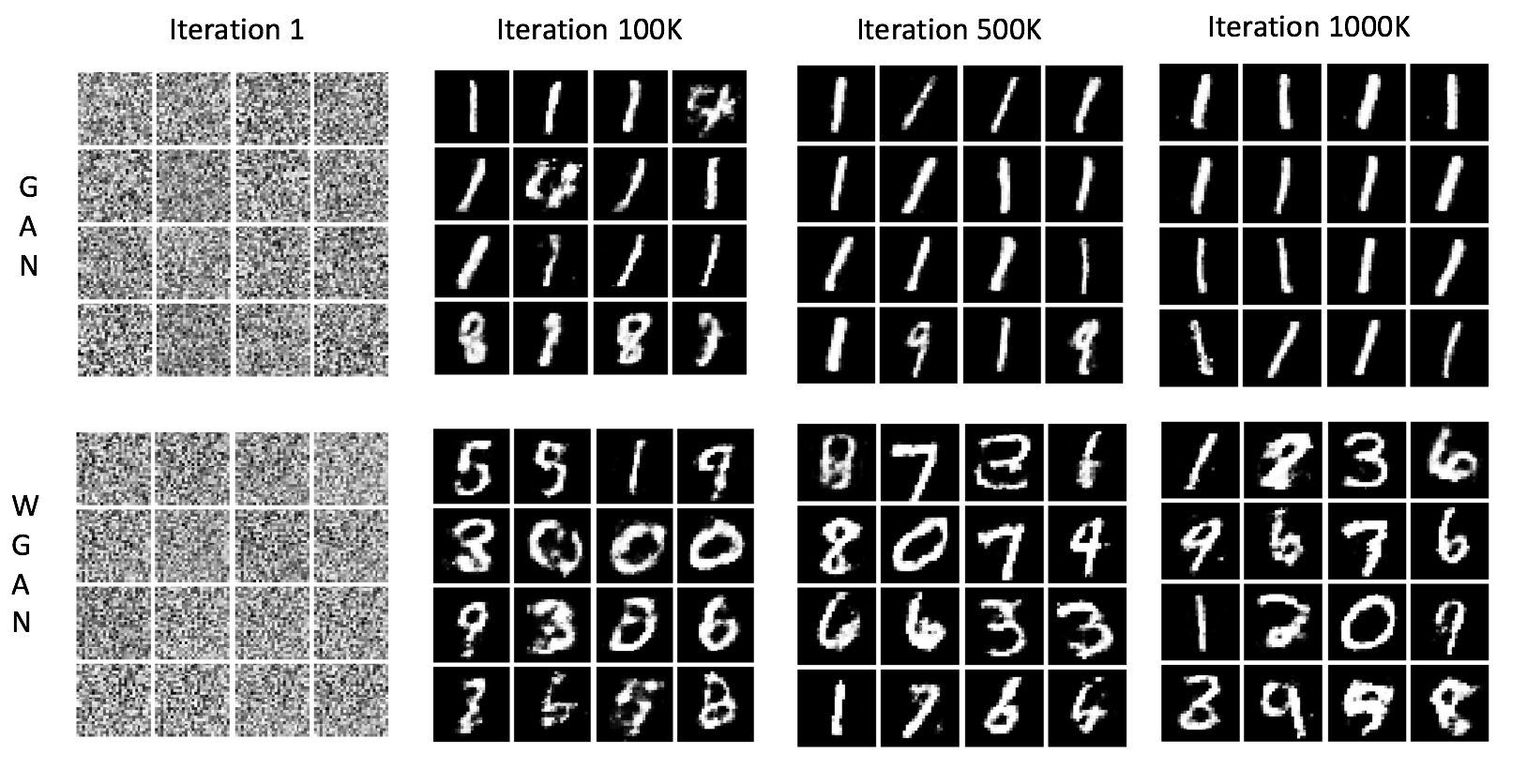}}
\caption{Generated images by GAN and WGAN models trained on MNIST after $1$,$100k$,$500k$,$1000k$ iterations.}
\label{ganwgan}
\end{center}
\vskip -0.2in
\end{figure*}

\begin{figure*}[ht]
\vskip 0.2in
\begin{center}
\centerline{\includegraphics[width=0.8\linewidth]{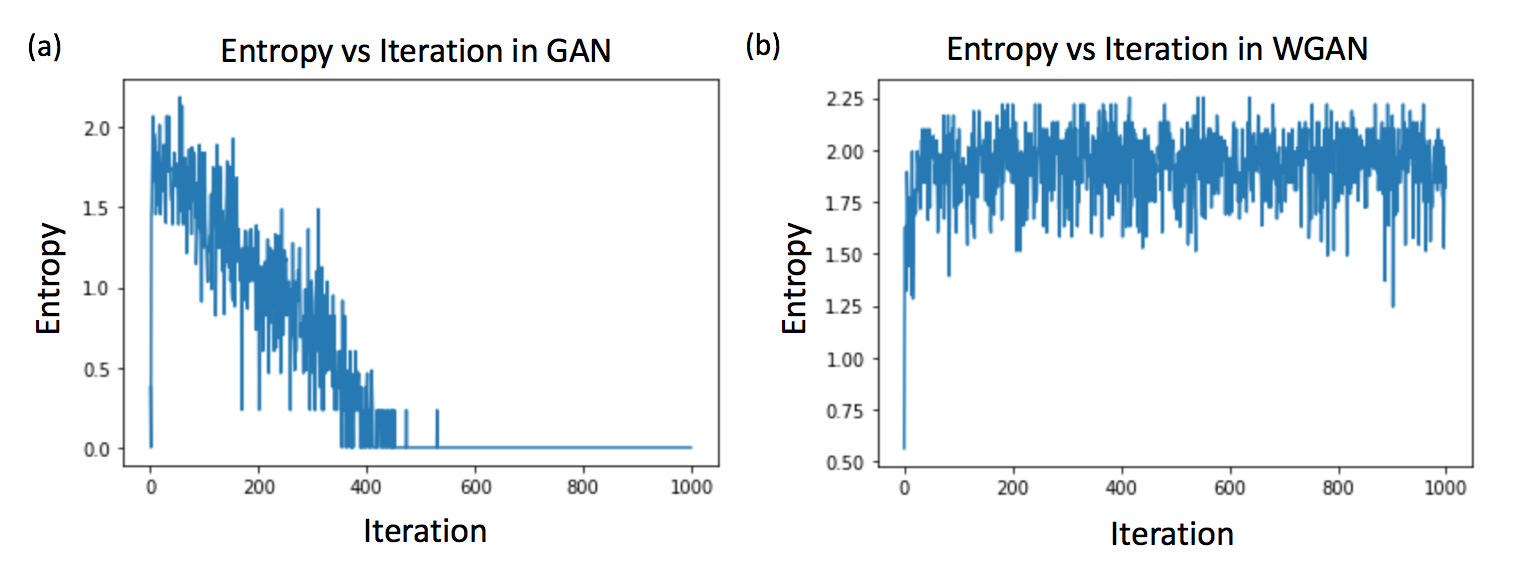}}
\caption{(a) The variation of entropy vs iterations in training process of GAN. (b) The variation of entropy vs iterations in training process of WGAN.}
\label{entropy}
\end{center}
\vskip -0.2in
\end{figure*} 
\begin{figure*}[ht]
\vskip 0.2in
\begin{center}
\centerline{\includegraphics[width=1\linewidth]{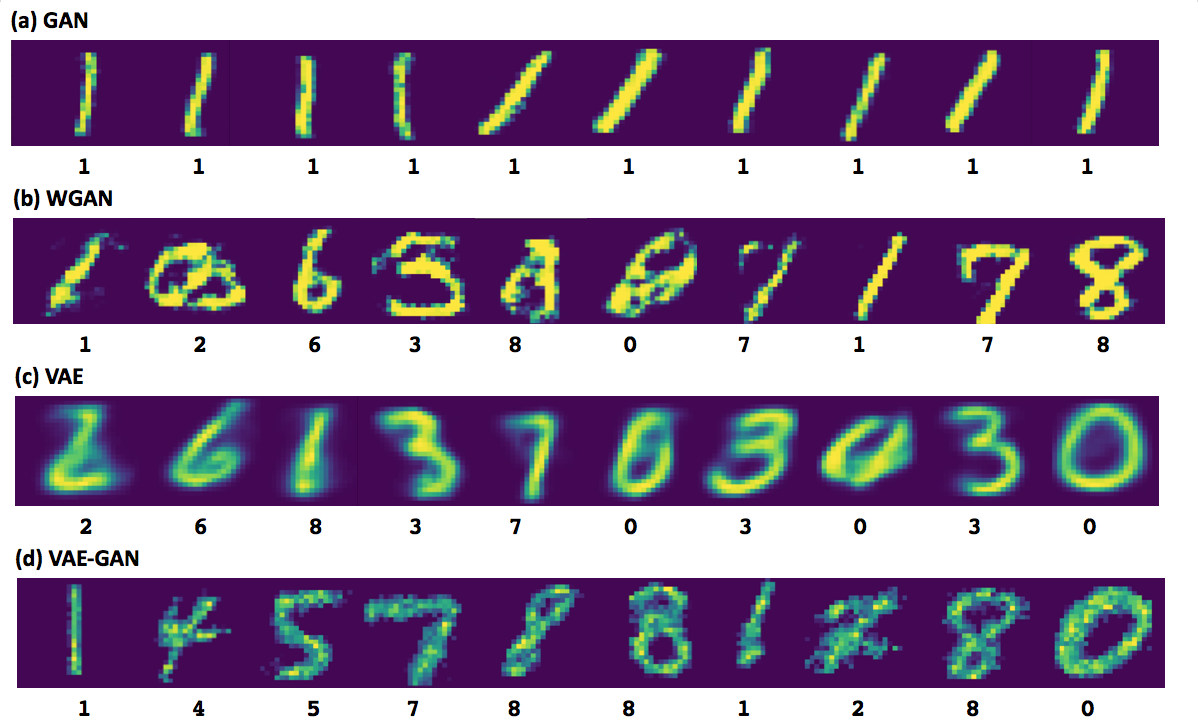}}
\caption{10 handwritten images sampled from model (a)GAN (b)WGAN (c)VAE (d)VAE-GAN, the labels under each row of images are predicted by a well-trained two-layer fully-connected neural network classifier.}
\label{imagequality}
\end{center}
\vskip -0.2in
\end{figure*} 

\begin{figure}[ht]
\vskip 0.2in
\begin{center}
\centerline{\includegraphics[width=\linewidth]{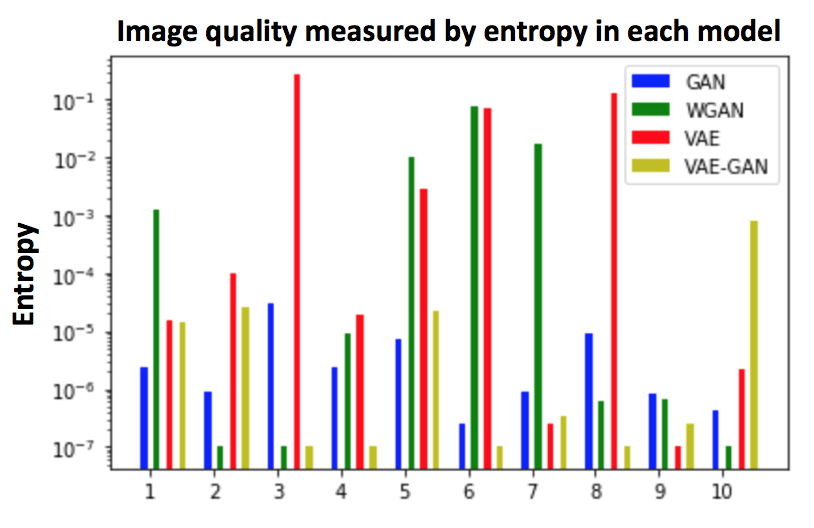}}
\caption{The single image entropy of each handwritten image in Fig.\ref{imagequality} computed from the prediction probability of a pre-trained classifier. A low entropy means that the image quality is high as it suggests that the classifier recognizes the image easily. The x axis label represents the index of each image generated from different models.}
\label{ime}
\end{center}
\vskip -0.2in
\end{figure}

\section{Experiments}
\subsection{Settings}
%Please describe the network structure and all metrics that we used.
In this section, we experiment with GAN, WGAN, VAE and VAE-GAN to quantitatively analyze the performance of mitigating mode collapse based on MNIST dataset. 

Firstly, we implemented all the generative adversarial models discussed so far with Tensorflow using the same fully connected neural networks for both the generator and the discriminator. In each model, the discriminative network is composed of five layers, where the size of input layer is 784, corresponding to the size of each 28$\times$28 handwritten image, and the size of output layer is 10$\times$10. The size of the layers between the output and the input are 392, 196, 98; the activation function in each layer between is ReLu function, and we use sigmoid function for the output layer. The generative network has exactly the same structure with the discriminative network, with the layers in the reverse order. The VAE model has analogous encoder network and decoder network structure as the discriminative and generative network mentioned above. For VAE-GAN, in addition to the same encoder and decoder network as in VAE model, an additional discriminative network are added to the end of decoder and another input layer of MNIST dataset. Differences between each model are mainly the definitions of loss functions and methods to realize gradient descent and decrease loss. The training process of GAN is to decrease loss from discriminator and generator defined as Eq.\ref{gan_loss2}. The process of WGAN is to minimize the loss shown in Eq.\ref{wgan}, which is implicitly minimizing the Wasserstein distance between the generative distribution and the data distribution. For VAE, the training process is to reduce the mean squared error between itself and the target and the KL divergence between the encoded latent variable and standard normal distribution, as defined in Eq.\ref{vae}. For VAE-GAN, the final loss function combined two parts, the loss generated in VAE part as Eq.\ref{vae} and loss generated in GAN part as Eq.\ref{vae-ganloss}, are finally used to generate the synthetic images.

\subsection{Entropy}
In order to provide a quantitative analysis, we used entropy of the synthetic data distribution to measure the severity of mode collapse. A classifier composed of 2-layer neural networks is firstly trained on the full MNIST dataset, which achieved $95.4\%$ accuracy on the test dataset. Then we used the classifier to recognize the handwritten digits generated from GAN and WGAN and calculated the entropy of the generative distribution for each training iteration. Let $p_i$ represent the probability of each digit $i$ sampled from the generative network at each iteration. We use the empirical estimator 
$$p_i = \frac{\text{number images classified as digit i}}{\text{number of images generated in total}}$$
Then the entropy of the generator can be estimated as
$$Entropy = -\sum_{i}^{10} p_i\times log(p_i)$$
When mode collapse happens, the entropy will keep decreasing. For example, we show the training process of GAN and WGAN, shown in Fig. \ref{ganwgan}. After 1000k iterations, the final images sampled from GAN only contain digit 1, which indicates that it is prone to mode collapse. In contrast, for WGAN, the final result is composed of various different digits, the mode collapse issue is mitigated significantly. The entropy shown in Fig. \ref{entropy} of each iteration decreases rapidly, and reaches zero after 480k iterations, which represents complete mode collapse. In contrast, the entropy approaches a relatively steady value in the training process of WGAN.

Furthermore, we also compared the entropy of the output in the last iteration from GAN, WGAN, VAE, VAE-GAN, shown in Table.1. Only the entropy of the output in GAN is zero, and the results in the other models are all around 2.27, which indicates that all the modes are preserved. In other words, no mode collapse.

\begin{table}[tbp]
\centering  % 
\begin{tabular}{lccccc}  % {lccc} left-l,right-r,center-c
\hline
Model &GAN &WGAN &VAE &VAE-GAN \\ 
Entropy &0.0 &2.280 &2.266 &2.263 \\ 
\hline
\end{tabular}
\caption{Entropy of images generated from GAN, WGAN, VAE, VAE-GAN.}
\end{table}

\begin{figure*}[htpb]
\vskip 0.2in
\begin{center}
\centerline{\includegraphics[width=0.8\linewidth]{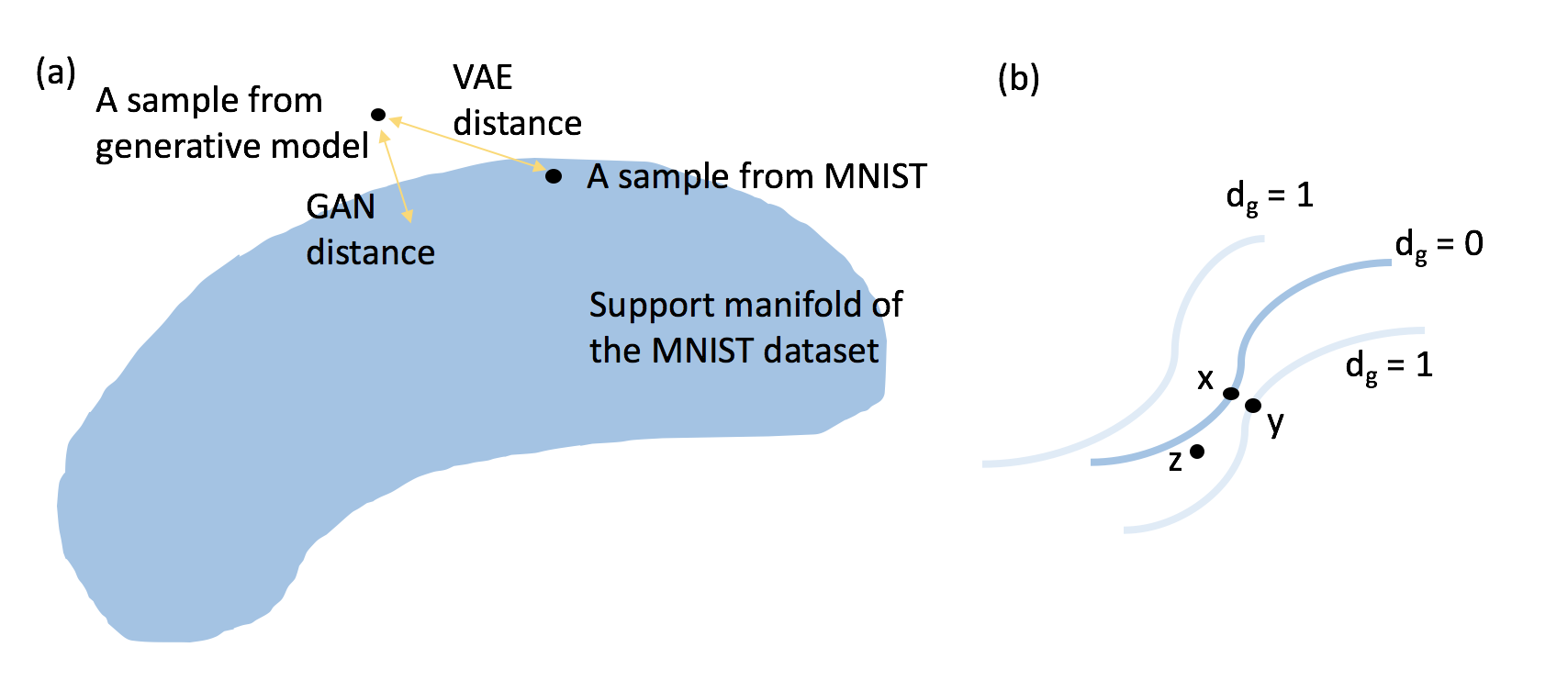}}
\caption{(a) The blue surface represents the low dimension structure of the MNIST data embedded in $\mathcal{R}^{28 \times 28}$. (b) Assume that the center blue line represent the support of the true distribution. The light colored lines are points in the high dimensional space that have distance 1 to the true distribution, measured by the discriminator in GAN model. Points x, y, z represent images in  $\mathcal{R}^{28 \times 28}$. }
\label{manifold}
\end{center}
\vskip -0.2in
\end{figure*} 

\subsection{Image quality}
We further compare the image quality from each well-trained model, and sample 10 images from the output layer of GAN, WGAN, VAE and VAE-GAN in the last iteration, shown in Fig. \ref{imagequality}. The labels under each row of images are predicted by a well-trained two-layer fully-connected neural network classifier. Images sampled from GAN only contain digit 1 due to mode collapse, from which other models do not suffer. However, the result of WGAN has a relatively low quality compared with the the original dataset: some images are hard to recognize with naked eyes. The results of VAE also do not achieve a satisfactory quality due to blurs. The images in VAE-GAN have the best performance in terms of image quality among those models. \\
In order to provide a quantitative study for image quality, we calculate the entropy of each single generated image. More specifically, for the $k-th$ image, we  compute $\sum_i p_{i,k}\log(p_{i,k})$, where $p_{i,k}$ is the predicted probability that image k belongs to class i. The prediction is done using a pre-trained classifier. A high-quality output is likely to be classified with high confidence and hence will have a low entropy. The result is shown in Fig. \ref{ime}. The entropy of VAE-GAN is lower than that of the other models, while the entropy of several images generated from VAE and WGAN are high due to their low quality. The entropy of GAN is low due to the fact that mode collapse enables the model to be easily optimized.

\begin{figure*}[hptb]
\vskip 0.2in
\begin{center}
\centerline{\includegraphics[width=0.9\linewidth]{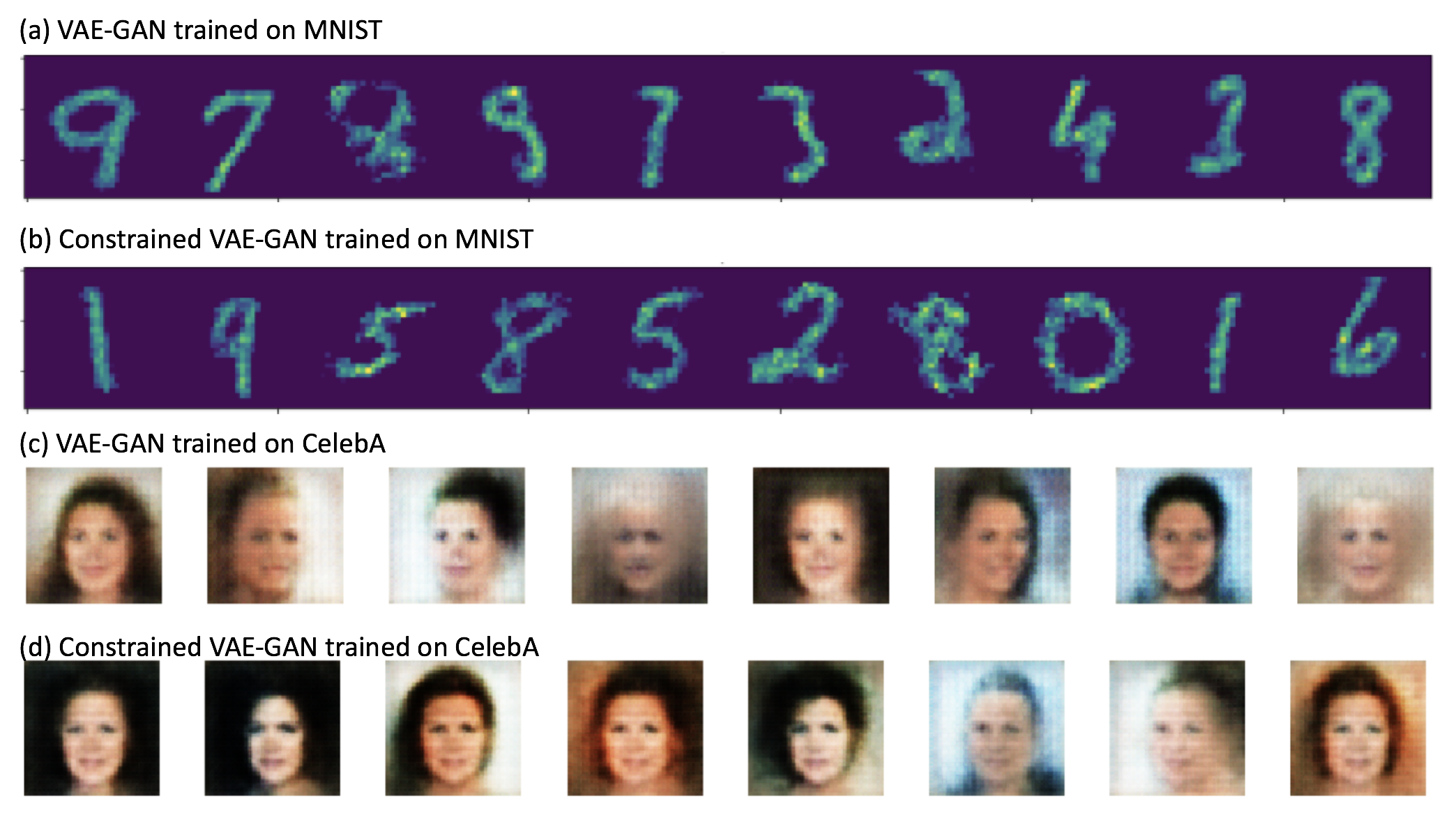}}
\caption{(a) Images sampled from VAE-GAN after training on the MNIST dataset. (b) Images sampled from our proposed model after training on the MNIST dataset. (c) Generated images from VAE-GAN after training 25 epochs on the CelebA dataset. (d) Generated images from constrained VAE-GAN after training 25 epochs on the CelebA dataset.}
\label{c-vaegan}
\end{center}
\vskip -0.2in
\end{figure*} 

\section{Our proposed model}
In this section, we wish to provide an argument on why VAE models and GAN models generate very different styles of samples. Based on this argument, we explain how the VAE-GAN models can be potentially improved. We then implement this idea and test the model on MNIST dataset.  

\subsection{Motivation}
From our experiment results in Section 3, we notice that each model has its own advantage over the rest even though they share the same network structure. The original GAN model proposed in \cite{gan} produces images of the highest quality. However, since its loss function over-emphasizes its ability to fool the discriminator, the model only produces simple images and suffers from mode collapse. The WGAN model \cite{wgan} generates images that resemble digits with various shapes and hence solves the mode collapsing problem. However, some samples clearly belong to none of the 10 classes from a human's perspective.  VAE models \cite{vae}, on the other hand, produces pretty images without mode collapsing. However, these images are blurry and can be easily distinguished from samples from the original dataset. The VAE-GAN model \cite{vaegan} attempts to get the best of both, but ends up producing samples similar to the ones from WGAN. However, it has the same advantage of getting latent vector distribution from an input image as the VAE models do.\\
We propose the following explanation for the behaviors described above. Assume that the true distribution of MNIST dataset lies on a low dimensional manifold as shown in Fig. \ref{manifold} and an instance is sampled from the deep generative model. We think of the loss functions for VAE models and GAN models as distance functions. The VAE distance measures its distance to the original input image. In particular, when the conditional distribution is Gaussian in VAE models, the decoder's loss is proportional to Euclidean distance. As a result, original VAE models generate images close to the input data points in Euclidean distance. Hence, these images have similar shapes as the true data but admit small deviations in pixel values and are blurry.  On the other hand, GAN distance measures the fake point's distance to the manifold, and equals zero as long as the point is on the manifold. Therefore, the GAN models produce points that are very close to the true distribution's support. However, without regularization based on the true data points, these generated images may not span the entire manifold. Furthermore, the discriminator learned may not properly compute the distance due to the difficulty of non-convex optimization.\\
As explained in section 2.4, the VAE-GAN model tries to optimize both loss functions at the same time. The loss for the decoder and encoder can be written into two components as follows,
$$ L_{vae-gan}(x)  =  L_{gan}(x) + L_{vae}$$
Yet, the result suggests that the GAN loss might dominate the other since it strengthens over iterations. We wish to design a model that allows VAE loss to take effect. Hence, we propose a constrained loss 
$$ \min_x L_{vae}(x), \ s.t.\ L_{gan}(x) \le d $$
The justification for this model is illustrated in Fig. \ref{manifold}. If point x is the input image.  Then point z has a lower loss in the original VAE-GAN model due to its low GAN loss, but y would have a lower loss in our constrained model, since it is in the feasible region and has a lower VAE loss. Solving this constrained problem allows the model to focus on imitating the shape of the original data samples as long as the image quality can almost fool the discriminator. In order to solve this constrained problem, we rewrite it in its Lagrangian form
$$ \mathcal{L}(\lambda, x) = L_{vae}(x) + \lambda(L_{gan}(x) - d)  \ s.t.\ \lambda \ge 0 $$
By KKT conditions, we can find local optima by solving $\mathcal{L}(\lambda^*, x)$, where $\lambda^* $ that maximize $\mathcal{L}(\lambda, x)$. It is straightforward to check that $\lambda^* = 0$ when $L_{gan}(x) \le d$, and  $\lambda^* = \infty$ when $L_{gan}(x) > d$. As an approximation, we solve the following problem for a fixed $\lambda > 0$,
$$ \min_x [L_{vae}(x) + \lambda \max\{L_{gan}(x)-d, 0\}] $$
The sub-gradient for this loss function can be easily computed and we can train the neural network with back-propagation as usual.

\subsection{Experiment}

The only difference between our model and the VAE-GAN model is that it uses a nonlinear combination of the loss from the VAE model and the loss from the discriminator. Hence, to control all other factors, we only changed the loss function in our original code for training VAE-GAN models, and ran stochastic gradient descent algorithm for the same number of iterations. \\

\begin{figure}[htbp]
\vskip 0.2in
\begin{center}
\centerline{\includegraphics[width=\linewidth]{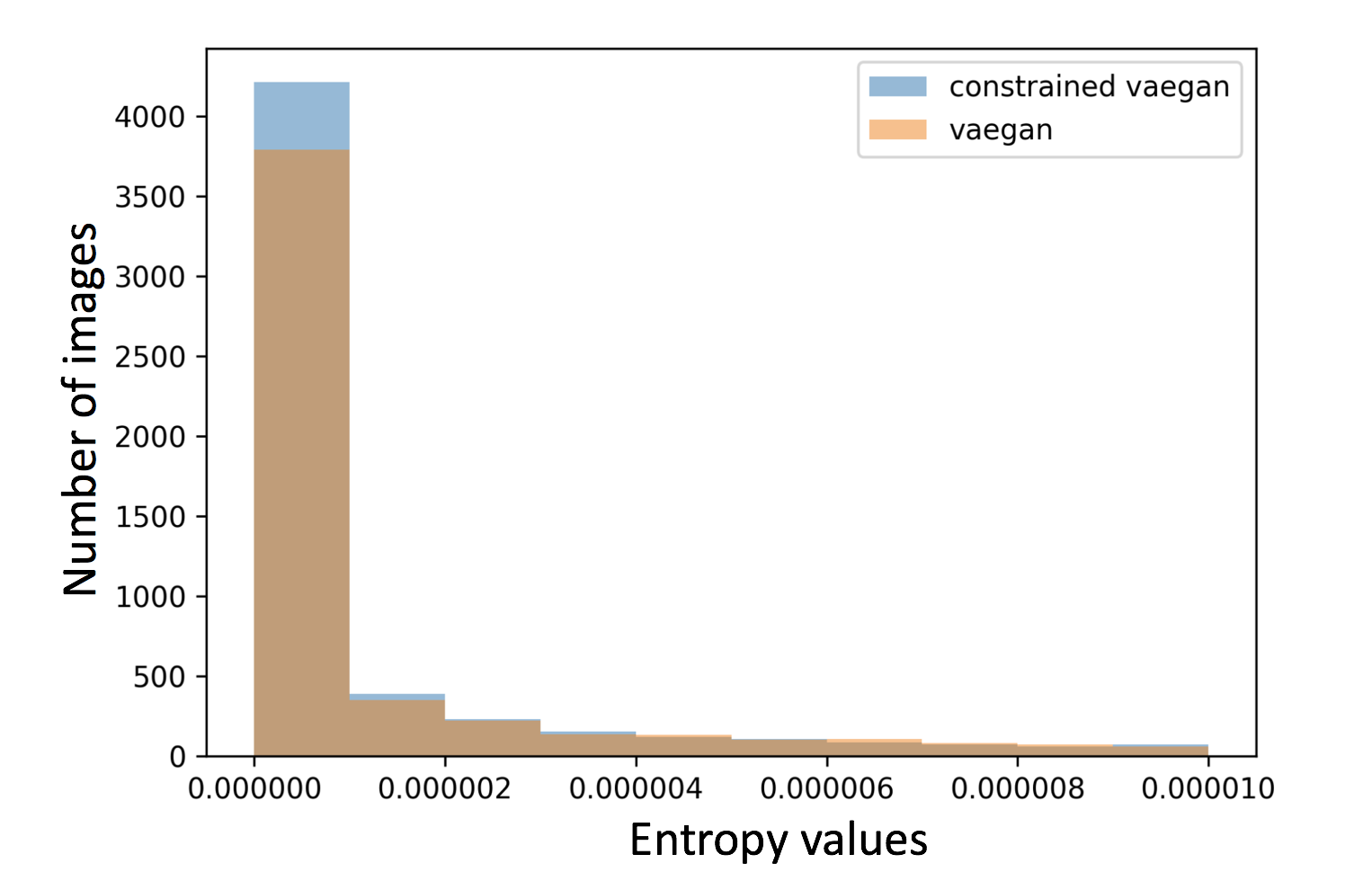}}
\caption{We sample 10k images from both constrained vaegan and vaegan models. We then use a pretrained classifier to classify each single image. For every image, we can compute its own entropy using predicted probability for each class. A high quality image can be recognized easily and hence should have low entropy. }
\label{cvaegan-entropy}
\end{center}
\vskip -0.2in
\end{figure}

First, we run both models on MNIST dataset. Both models use the same 5-layer fully connected neural networks as introduced in Section 3. The sampled random images are shown in fig. \ref{c-vaegan}(a)(b). Neither of the two models has a dominating performance, but our proposed one seems to have more stable image quality. This is verified in fig. \ref{cvaegan-entropy}. The entropy here is defined the same way as the entropy in fig. \ref{ime}. Low entropy is associated with high image quality. We notice that our proposed algorithm has higher concentration of low entropy images compared to the original VAEGAN model.\\
Then we tested both models on the CelebA dataset with convolutional neural networks. Our network architecture has three convolution layers and is the same as the original paper in \cite{vaegan}. The original code runs the training process for about 50 epochs, but due to our limited computation resource, we have to terminate the process at epoch 25 after training for an entire week. Some preliminary results are shown in fig. \ref{c-vaegan}(c)(d). We are not able to draw any interesting conclusion since the network has not converged.

\section{Discussion}
There are a few problems unsolved due to our limited time and computation resource. First, MNIST is a dataset with a simple structure. Therefore, the conclusions we draw based on MNIST experiments may not generalize to more complicated data. Even though we attempt to train some convolutional neural networks(CNN) on the CelebA data set, the lack of GPU access forces us to terminate the experiment before it finishes training. Therefore, it would be interesting to check if our proposed model can have greater improvement when the manifold in high dimensional space is more complicated. Second, the quality of the images generated in our experiments are relatively low compared to results in more recent literatures. This results from the disadvantage of fully connected neural network compared with CNN in learning image structures. We suspect that our WGAN model generates low quality images because the network is not powerful enough. Again, we do not have enough resources to conduct experiments that require convolution operations.\\
In the results shown, we tried to make fair comparison by using the same network architecture in all of our models. We trained VAE-GAN and constrained VAE-GAN with the same parameters for the same number of iterations. Other models are trained until the image quality stabilizes. From these results, we may conclude that our explanation in in section 4 aligns with the experiments in Section 3. However, the experiment comparing VAE-GAN and constrained VAE-GAN shows little difference, and more efforts are needed before we could get stronger evidence to support our claim.

\bibliography{example_paper}

\begin{thebibliography}{12}
\providecommand{\natexlab}[1]{#1}
\providecommand{\url}[1]{\texttt{#1}}
\expandafter\ifx\csname urlstyle\endcsname\relax
  \providecommand{\doi}[1]{doi: #1}\else
  \providecommand{\doi}{doi: \begingroup \urlstyle{rm}\Url}\fi

\bibitem[Andrieu et~al.(2003)Andrieu, De~Freitas, Doucet, and Jordan]{mcmc}
Andrieu, Christophe, De~Freitas, Nando, Doucet, Arnaud, and Jordan, Michael~I.
\newblock An introduction to mcmc for machine learning.
\newblock \emph{Machine learning}, 50\penalty0 (1-2):\penalty0 5--43, 2003.

\bibitem[Arjovsky et~al.(2017)Arjovsky, Chintala, and Bottou]{wgan}
Arjovsky, Martin, Chintala, Soumith, and Bottou, L{\'e}on.
\newblock Wasserstein gan.
\newblock \emph{arXiv preprint arXiv:1701.07875}, 2017.

\bibitem[Blei et~al.(2003)Blei, Ng, and Jordan]{lda}
Blei, David~M, Ng, Andrew~Y, and Jordan, Michael~I.
\newblock Latent dirichlet allocation.
\newblock \emph{Journal of machine Learning research}, 3\penalty0
  (Jan):\penalty0 993--1022, 2003.

\bibitem[Goodfellow et~al.(2014)Goodfellow, Pouget-Abadie, Mirza, Xu,
  Warde-Farley, Ozair, Courville, and Bengio]{gan}
Goodfellow, Ian, Pouget-Abadie, Jean, Mirza, Mehdi, Xu, Bing, Warde-Farley,
  David, Ozair, Sherjil, Courville, Aaron, and Bengio, Yoshua.
\newblock Generative adversarial nets.
\newblock In \emph{Advances in neural information processing systems}, pp.\
  2672--2680, 2014.

\bibitem[Kingma \& Welling(2013)Kingma and Welling]{vae}
Kingma, Diederik~P and Welling, Max.
\newblock Auto-encoding variational bayes.
\newblock \emph{arXiv preprint arXiv:1312.6114}, 2013.

\bibitem[Larsen et~al.(2015)Larsen, S{\o}nderby, Larochelle, and
  Winther]{vaegan}
Larsen, Anders Boesen~Lindbo, S{\o}nderby, S{\o}ren~Kaae, Larochelle, Hugo, and
  Winther, Ole.
\newblock Autoencoding beyond pixels using a learned similarity metric.
\newblock \emph{arXiv preprint arXiv:1512.09300}, 2015.

\bibitem[Loaiza-Ganem et~al.(2017)Loaiza-Ganem, Gao, and Cunningham]{entropy}
Loaiza-Ganem, Gabriel, Gao, Yuanjun, and Cunningham, John~P.
\newblock Maximum entropy flow networks.
\newblock \emph{arXiv preprint arXiv:1701.03504}, 2017.

\bibitem[Mirza \& Osindero(2014)Mirza and Osindero]{cgan}
Mirza, Mehdi and Osindero, Simon.
\newblock Conditional generative adversarial nets.
\newblock \emph{arXiv preprint arXiv:1411.1784}, 2014.

\bibitem[Radford et~al.(2015)Radford, Metz, and Chintala]{dcgan}
Radford, Alec, Metz, Luke, and Chintala, Soumith.
\newblock Unsupervised representation learning with deep convolutional
  generative adversarial networks.
\newblock \emph{arXiv preprint arXiv:1511.06434}, 2015.

\bibitem[Ranganath et~al.(2016)Ranganath, Tran, and Blei]{hierarchy}
Ranganath, Rajesh, Tran, Dustin, and Blei, David.
\newblock Hierarchical variational models.
\newblock In \emph{International Conference on Machine Learning}, pp.\
  324--333, 2016.

\bibitem[Rezende \& Mohamed(2015)Rezende and Mohamed]{networkflow}
Rezende, Danilo~Jimenez and Mohamed, Shakir.
\newblock Variational inference with normalizing flows.
\newblock \emph{arXiv preprint arXiv:1505.05770}, 2015.

\bibitem[Wainwright et~al.(2008)Wainwright, Jordan, et~al.]{inference}
Wainwright, Martin~J, Jordan, Michael~I, et~al.
\newblock Graphical models, exponential families, and variational inference.
\newblock \emph{Foundations and Trends{\textregistered} in Machine Learning},
  1\penalty0 (1--2):\penalty0 1--305, 2008.

\end{thebibliography}
\bibliographystyle{icml2017}

\end{document}